# Anatomy of a Robotaxi Crash:
# Lessons from the Cruise Pedestrian Dragging Mishap

Philip Koopman[0000-0003-1658-2386]

Carnegie Mellon University, Pittsburgh PA, USA
koopman@cmu.edu

**Abstract.** An October 2023 crash between a GM Cruise robotaxi and a pedestrian in San Francisco resulted not only in a severe injury, but also dramatic upheaval at that company that will likely have lasting effects throughout the industry. Issues stem not just from the loss events themselves, but also from how Cruise mishandled dealing with their robotaxi dragging a pedestrian under the vehicle after the initial post-crash stop. External investigation reports provide raw material describing the incident and critique the company's response from a regulatory point of view, but exclude safety engineering recommendations from scope. We highlight specific facts and relationships among events by tying together different pieces of the external report material. We then explore safety lessons that might be learned related to: recognizing and responding to nearby mishaps, building an accurate world model of a post-collision scenario, the inadequacy of a so-called "minimal risk condition" strategy in complex situations, poor organizational discipline in responding to a mishap, overly aggressive post-collision automation choices that made a bad situation worse, and a reluctance to admit to a mishap causing much worse organizational harm downstream.

**Keywords:** Autonomous vehicles, robotaxi crash, regulation, safety lessons

## 1    Introduction

On October 2, 2023, a Cruise robotaxi Autonomous Vehicle (AV) in San Francisco struck and then – as part of a subsequent maneuver – dragged a pedestrian under the vehicle as part of a complex mishap scenario. A different, human-driven vehicle struck the pedestrian first. Cruise failed to proactively disclose the pedestrian dragging portion of the mishap. Many Cruise leaders were sacked, followed by a 24% workforce cut.

We examine the events of the mishap and the aftermath based on information made public in an external investigation report commissioned by Cruise [6]. That report is a single document file that contains two stand-alone parts with independent page numbering. The majority of references in this paper are to these two documents, with the tag for the report as indicated and a relevant report page number:

- QER: The primary report by Quinn Emanuel Trial Lawyers, with an emphasis on resolving questions of regulatory compliance and potential culpability. Much of its content is about who knew what, and who said what to whom, when. (109 pages)
- EXPR: A redacted technical root cause analysis for the mishap events. (86 pages)

The QER and EXPR state safety engineering and operational safety practices are out of their scope (QER 6, EXPR 13). However, they contain enough information to reconstruct a sequence of events that yields safety insights beyond what might be evident from a straightforward reading of the narrative and conclusions of those reports. A primary goal of this paper is to present a clarified view of those events.

At a high level, the QER portrays the crash itself as follows (QER 9-10). A pedestrian crossing against a "Do Not Walk" pedestrian signal stopped in traffic, and was first hit by a human driver in a different lane, who fled the scene. The pedestrian was "launched" into the path of the Cruise robotaxi AV by that first impact. The AV braked hard but was unable to avoid hitting the pedestrian. The subsequent pedestrian dragging was due to the AV doing something designed to enhance safety (achieving a so-called Minimal Risk Condition[1]) that went wrong due to the AV not detecting the pedestrian under the vehicle. None of this would have happened if a hit-and-run human driver of an adjacent vehicle had not hit the pedestrian first (QER 1,10).

The QER portrays the crash response and regulatory interfaces as due to "poor leadership, mistakes in judgment, lack of coordination, an 'us versus them' mentality with regulators, and a fundamental misapprehension of Cruise's obligations of accountability and transparency to the government and the public" (QER 7). The QER calls the ultimate regulatory operational suspension order "a direct result of a proverbial self-inflicted wound" due to mishandled interaction with regulators (QER 7).

A closer look at the material in the reports, however, reveals there is much to understand beyond that narrative. Indeed, the human driver in the other vehicle acted badly, and since-sacked Cruise management mishandled the crisis. However, there is much more texture to the situation, including significant technical shortcomings in the AV's design as well as significant room for improvement in operational safety procedures.

The AV potentially violated a California road rule by accelerating toward a pedestrian in a crosswalk. The AV had trajectory information on the human-driven vehicle and pedestrian a few feet away, but failed to recognize it as a pending collision. The AV could have stopped much sooner, even if only reacting to the pedestrian intrusion into the AV's lane, avoiding or mitigating initial impact injuries. The pedestrian was not thrown completely onto the ground at impact, but rather was at least partly on the hood of the AV before being run over. Multiple AV sensors indicated the pedestrian's presence under the vehicle, but the AV failed to recognize the scenario. The AV proceeded with a post-crash maneuver essentially instantly without giving a remote assistance team time to assess the situation. And, the contractor-staffed remote assistance team seemed to know of the dragging essentially immediately, but the Cruise crisis response team had to figure that out for themselves early the next morning.

We agree with the QER that cultural issues contributed to the regulatory mess created. But, we go further and say that fundamental changes are needed in the technical systems, operational procedures, and crisis response approach to address deep safety concerns beyond just the compliance topics within the scope of the QER.

---

[1] Despite the name, an MRC does not guarantee global minimal risk, but rather is merely a "stable stopped" condition "to reduce the risk of crash." The risk need not be objectively "minimal", and the defining SAE J3016 recommended practice disclaims safety scope [7].

## 2      Background

### 2.1     Terminology

Below are some key terms, with Cruise-specific terms defined based on the QER.
- CIRT: Cruise Incident Response team: Cruise employees in San Francisco.
- CMT: Crisis Management Team: Cruise employees in San Francisco. This seems to be a follow-on to the CIRT, although the relationship is not explained in the QER.
- CPUC: California Public Utilities Commission: state government regulator of paid commercial transport permits in California.
- CRAC: Cruise Remote Assistance Center: contractors in Arizona.
- Cruise: a subsidiary of General Motors that is in the robotaxi business. Cruise is headquartered in San Francisco, with some relevant staff working remotely.
- DSS: Driverless Support Specialist contractors who provided on-scene support.
- DMV: California Department of Motor Vehicles: state government safety regulator.
- NHTSA: National Highway Traffic Safety Administration: US Federal government vehicle equipment safety regulator.
- City: Generic term encompassing San Francisco Metropolitan Transportation Authority (SF MTA), San Francisco Fire Department (SFFD), and the San Francisco Police Department (SFPD).

### 2.2     Crash context and overview

Cruise had been previously granted permission to operate without a safety driver by DMV, and granted permission to charge for robotaxi rides by CPUC. Numerous on-road incidents had occurred that established tension between Cruise and the City, especially involving interference with emergency responders and public road congestion. The CPUC public hearings were particularly contentious. A collision with a fire truck that resulted in an AV passenger injury occurred a week after the most recent CPUC approval, after which DMV requested that Cruise reduce its operational tempo [4].

There was no human driver in the mishap AV, nor was there a continuous real-time remote safety supervisor. The mishap AV had no passengers on board. The CRAC operations center located in Arizona (perhaps 700-800 miles distant) oversaw remote operations. The DSS support team provided local on-road support in San Francisco, for example, to recover disabled vehicles. Engineering activities and the CIRT/CMT crisis response teams were centered in San Francisco, but drew upon staff in other states.

A mishap overview starts at EXPR 14. The mishap occurred starting at 9:29 PM US PT on October 2, 2023. Environmental factors are not mentioned as contributing issues, although at that time it would have been dark with some streetlight illumination. The speed limit was 25 mph (EXPR 18), with a maximum AV mishap speed of 19.1 mph.

Figure 1 shows a simplified diagram of the mishap geometry. At the start of the mishap sequence, two vehicles were stopped at a red traffic light next to each other at an intersection on a 4-lane surface road: the Cruise AV on the curb lane, and a human-driven Nissan in the medial lane, adjacent to the two-way street center-line.

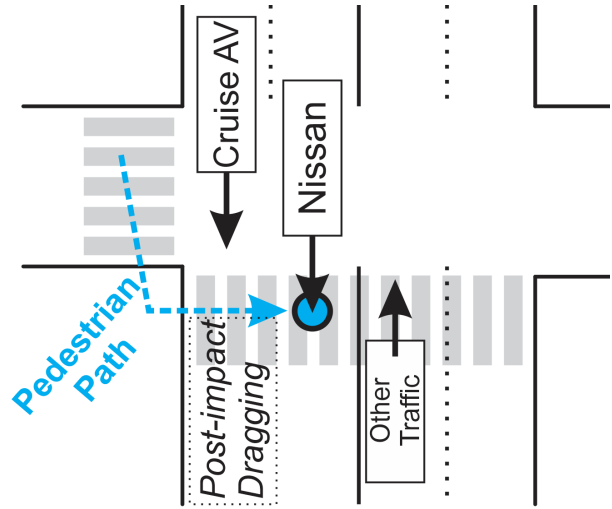

*Figure 1. Simplified diagram of mishap; not to scale.*

After the vehicles' light had changed to green, a pedestrian entered the crosswalk on the far side of the intersection. The pedestrian first crossed the AV's lane, and then the Nissan's lane. Both vehicles accelerated straight through the intersection toward the far crosswalk as the pedestrian traversed their paths. The pedestrian stopped in front of the Nissan, with her path blocked by oncoming traffic in the other direction. The pedestrian was likely distracted from the approaching Nissan ("did not look left") while attempting to signal opposing traffic to permit continuing the crossing (interpretation of description at EXPR 47). The Nissan struck the pedestrian in its lane, braking only after impact.

After an impact interaction with the Nissan, the pedestrian separated from the Nissan and entered the AV's travel lane. The AV started braking just before impact, hitting the pedestrian at close to its maximum pre-impact speed, coming to a quick stop. The AV's front wheel ran over the pedestrian, leaving the pedestrian under the AV.

A split-second after coming to a stop, the AV initiated a pullover maneuver to achieve a so-called "Minimal Risk Condition" (MRC), incorrectly assessing the situation as a side impact rather than a run-over scenario (EXPR 16). CRAC remote operators connected with the vehicle while it was partway through this maneuver.

The pedestrian was entrapped under the vehicle, and was dragged along the pavement for 20 feet at speeds up to 7.7 mph. The AV might have continued this dragging situation for up to 100 feet (QER 14). However, the AV recognized a vehicle motion anomaly and stopped the motion prematurely without recognizing a trapped pedestrian.

We consider details in three phases: the day of the mishap (crash and immediate response), the day after the mishap (post-crash response), and the longer-term regulatory interactions. The regulatory interactions lasted for several weeks, with Cruise ultimately shutting down their public robotaxi operations for an extended period. Each phase presents lessons to be learned not only by Cruise, but by any company testing or deploying automated vehicle technology on public roads.

All times given in this paper are in the US PT (local) time zone. Relative times are from the time of the initial AV impact with the pedestrian, rounded to the nearest time unit available in the source material. While the QER and EXPR have several timelines and additional information in various places, we integrate the timelines to show some relationships and considerations not readily apparent in the source material.

## 3   Crash details

The QER states "But for the human driver of the Nissan hitting the pedestrian, the October 2 Accident would not have occurred" (QER 10). While narrowly true, this is not the whole story. Adverse events not caused by a robotaxi can be expected, and to achieve acceptable safety it is a near certainty that an AV will need to be able to do something reasonable in response to the vast majority of such events. As a counterfactual example, if the pedestrian had seen the Nissan coming, she might have suddenly run back across the AV's path to reach the sidewalk. Given that the AV was tracking the pedestrian being caught by traffic in an adjacent lane, information regarding an emergent high-risk traffic situation was available to, but not recognized by, the AV.

Regardless of what might have been, more relevant to improving future safety are: What went right? What went wrong? And what lessons might be learned?

### 3.1   Crash timeline

The following timeline highlights points especially relevant for our discussion. More detailed timelines and time series data graphs are available, but require some integration on the part of the reader (QER 10 *et seq.*, 18; EXPR 43-45, 50, 54, 68, 83, 85). Figure 2 shows a portion of the velocity and acceleration graphs during the mishap sequence.

Items in this timeline are sourced from EXPR 44 unless otherwise noted.

- -38.3 sec: Pedestrian tracking begins. Pedestrian crosses street parallel to later vehicle motion, then turns left to cross in front of the vehicles on the far side of intersection. Pedestrian remains in clear sight of AV the entire time. Whether the AV tracked pedestrian gaze as attempting "eye contact" with AV or not before starting to cross at -7.9 seconds is not mentioned (EXPR 19, 48-49).
- -10 sec.: Traffic light changes; both vehicles are motionless.
- -9.2 sec: By this time, both Nissan and AV are accelerating straight toward an empty crosswalk on the far side of the intersection. The Nissan leads slightly.
- -7.9 sec: Pedestrian enters crosswalk. AV traveling approximately 5.5 mph (fig. 2).
- -7.7 sec: AV predicts pedestrian will cross AV's lane. AV continues accelerating.
- -5.3 sec: Pedestrian leaves AV travel lane after 2.6 seconds. AV traveling approximately 13.5 mph, and continues to accelerate (fig. 2).
- -4.8 sec: AV predicted paths for Nissan and Pedestrian "are consistent with a potential collision", but "the ADS did not consider the potential of a collision between the Nissan and the pedestrian" (EXPR 15, 50, 53, 77).
- -4.7 sec: Pedestrian stops/pauses in Nissan's travel lane, blocked by traffic in the opposing direction, and remains in crosswalk (EXPR 47).
- -4.6 sec: AV biases right in its lane, possibly due to the presence of the pedestrian in adjacent lane (EXPR 51).

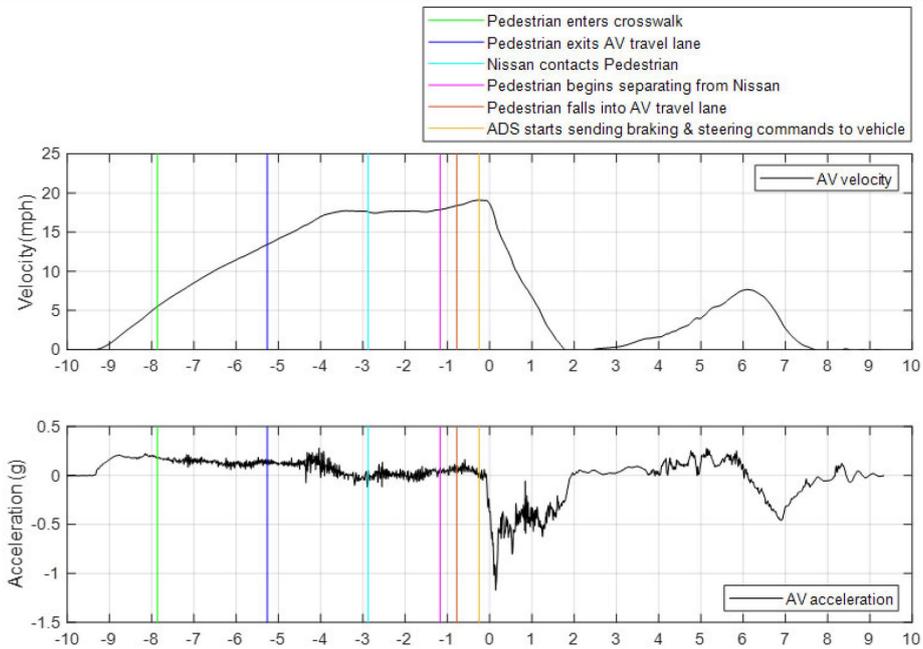

*Figure 2. AV velocity and acceleration profiles from EXPR 43.*

- -2.9 sec: Nissan impacts pedestrian at 21.7 mph without prior braking (EXPR 15, 52). AV is now at 17.6 mph (EXPR 53) and maintains this speed through the next second (EXPR 55). The pedestrian is moving at 2.6 mph at the time of collision, but the direction of motion is redacted (EXPR 51-52). The AV onboard camera captures a frame of that collision event, which was in view of the AV (EXPR 78).
- -2.0 sec: AV Pedestrian tracking is dropped. However, intermittent classification and tracking continues until -0.3 sec (EXPR 79).
- -1.5 sec (approx.): AV resumes gentle acceleration (Fig. 2)
- -1.17 sec: Visual separation of Pedestrian from Nissan. AV at 17.9 mph (EXPR 15).
- -1.07 sec: Nissan initiates hard braking (brake lights) (EXPR 78).
- -0.78 sec: Pedestrian enters AV's travel lane approximately 21.5 feet in front of AV. AV at 18.4 mph (EXPR 66).
- -0.78 sec: Initiating braking now would have completely avoided pedestrian contact, but AV did not brake, instead increasing speed slightly (EXPR 66, fig. 2).
- -0.41: imminent collision predicted by AV "collision checker".
- -0.3 sec: AV last correct classification as a pedestrian (EXPR 15, 79).
- Intermittent classification and tracking of pedestrian. However, AV detecting "occupied space", leading to steering and braking commands (EXPR 15, 79).
- -0.25 sec: AV sends steering and braking commands. AV at 19.1 mph (EXPR 15).
- -0.2 sec: Nissan has completely stopped[2]

---

[2] There is a small discrepancy with the timing information given, with possible values of ranging from -0.2 to +0.4 sec (EXPR 56-57).

- 0 sec: AV initial impact with pedestrian. AV has slowed by 0.5 mph to 18.6 mph (EXPR 15). "Front bumper first contacted pedestrian" (EXPR 15). Collision detection system incorrectly identifies a side impact (EXPR 16). CRAC saw the pedestrian initially on the AV hood in a time-delayed video (QER 21).
- 0.23 sec: AV left front wheel runs over pedestrian (EXPR 16, 81).
- 1.78 sec: AV braking achieves a zero net speed (EXPR 16) for a fractional second.
- 1.83 sec: AV begins post-impact acceleration toward MRC. AV starts dragging the pedestrian approximately 20 feet at speeds of up to 7.7 mph (EXPR 16, 82-83).
- "within seconds" after impact, AV sends a 3-second collision video to CRAC that includes just the collision event (QER at 18).
- 3.8 sec: Left front wheel spins at perhaps 20 mph in a "traction control event" due to entrapped pedestrian "physically resisting the motion of the vehicle" (EXPR 16, 84, figure 64).
- Within 5 sec: CRAC connected to AV video and audio. CRAC saw "ped flung onto hood of AV. You could see and hear the bumps" and the AV "was already pulling over to the side." (QER 21, 61)
- 5.8 sec: Degraded mode entered due to wheel slip caused by pedestrian's legs (EXPR 16,84) is said to trigger an "immediate stop." In reality this mode triggers "gradually slowing to a stop" (EXPR 36), consistent with timeline. AV might instead have dragged the pedestrian up to 100 feet or one full block (QER at 34).
- 8.8 sec: Final point of rest for AV reached (EXPR 16, 84). Pedestrian is largely under rear of vehicle. Legs protrude from left rear, with tire on top of at least one pedestrian leg (QER 33). The pedestrian's feet and lower legs were visible in the wide-angle left side camera view throughout the event (EXPR 83). Legs were briefly detected but neither classified nor tracked after the collision (EXPR 16).

### 3.2    Crash analysis

**What went right:** An undifferentiated obstacle was seen in front of the AV, prompting an emergency stop due to a presumption that it might be a vulnerable road user (EXPR 17,32). The AV noticed an impaired ability to move accurately and entered degraded states accordingly. These degradations played a role in preventing the outcome from being worse, avoiding potentially dragging the pedestrian up to 100 feet.

**Accelerating into pedestrian:** The AV accelerated the entire time that it detected a pedestrian in a crosswalk in its travel lane, with speed more than doubling (8 mph increase) during the 2.6 seconds the pedestrian was in its travel lane. This was due to the AV's machine learning-based (EXPR 29) prediction that the pedestrian would be out of the way by the time the AV arrived (EXPR 44).

Accelerating into a pedestrian in a crosswalk is inconsistent with California Rules of the Road, which state: "(c) The driver of a vehicle approaching a pedestrian within any marked or unmarked crosswalk shall exercise all due care and *shall reduce the speed of the vehicle* or take any other action relating to the operation of the vehicle as necessary to safeguard the safety of the pedestrian." ([1] with emphasis added) Other California Vehicle Code sections relevant to the mishap are listed on QER 9 fn 12.

**Delay in braking:** The laws of physics did not preclude stopping in time to avoid AV/pedestrian impact. The AV had enough time to totally avoid impact if it had immediately responded to the pedestrian presence in its lane, even accounting for braking latency (EXPR 66). Even a delayed but earlier braking "would have potentially mitigated severity of the initial collision" (EXPR 66). Due to a slower-than-possible response, AV aggressive braking beyond 0.5g only occurred after impact time 0 (fig. 2).

**Brittle pedestrian position prediction:** The AV tracked the pedestrian coming to a stop in a traffic lane and to the point of being hit by an adjacent vehicle, which then braked aggressively. Any of these factors could have been treated as a red flag that a high-risk situation was developing, but were not.

The QER recounts statements regarding the impracticality of tracking a pedestrian in this particular impact incident, with staff using terms such as "unrealistic" and "insane hypothetical" (QER 66). However, at least some drivers would have braked or performed an evasive maneuver in this type of situation (EXPR 66).

**Vanishing pedestrian:** The AV acted in a way consistent with "forgetting" a pedestrian was in the vicinity. The AV did not stop because of a tracked pedestrian in its lane, but rather because there was occupied road space immediately in front of it that it assumed could be a vulnerable road user. It seems inappropriate to "forget" a pedestrian who has just been hit by a vehicle a few feet away. Moreover, there was sensor information showing the pedestrian, but the system was not up to the tracking challenge.

**Near-impact sensor self-occlusions.** Impact diagnosis depended on available object tracking information immediately prior to impact. However, this impact occurred in an area in which the pedestrian was substantially vehicle-self-occluded from lidar sensors, contributing to an incorrect assessment of the position of the pedestrian during and after impact (EXPR 80-81). Sensor self-occlusions near and under the vehicle contributed to AV misdiagnosis of the immediate pre-crash and post-crash situations.

**Moving with an entrapped pedestrian:** The vehicle had recent historical information available that a pedestrian was likely to have been struck and then lost to tracking. This situation should have required at least an override from CRAC before moving the vehicle further. Instead, the vehicle moved again, entirely on its own, approximately $1/20^{th}$ of a second after stopping. While a mapping error contributed to initiating movement (EXPR 17), any such system will need to be robust in the face of inevitable mapping errors. EXPR 17,76 admit: "After the AV contacted the pedestrian, an alert and attentive human driver would be aware that an impact of some sort had occurred and would not have continued driving without further investigating the situation."

**Functional insufficiencies.** The EXPR asserts no sources of hardware or software failure were identified (EXPR 14). This emphasizes the need to account for functional insufficiencies [3] beyond more narrowly defined implementation defects.

**The sensor-based safety narrative.** The AV industry safety narrative commonly features an emphasis on superior sensor coverage combined with superhuman reaction time. Yet in this mishap the pedestrian was not tracked accurately, and braking was not initiated particularly quickly for a computer driver. EXPR 17, 85 admit: "The AV's lack of anticipation of a potential future incursion of the pedestrian into its travel lane was a contributing factor to this incident."

Sensors and fast reactions only present the possibility of safety, which must be complemented by robust classification and tracking capabilities. The reasoning ability of the vehicle did not include consideration of a disappearing pedestrian who had just been hit. Despite multiple kinesthetic clues and AV camera imagery showing the legs of a pedestrian being dragged under the vehicle, the AV proceeded with its pullover maneuver, continuing until time 8.8 seconds.

**Safety comparison baseline.** The EXPR narrative switches apparent standards of comparison from a super-proficient robot (accelerating into a pedestrian against California Rules of the Road because it is confident the road will be clear when it arrives), to a naïve robot (failing to predict a collision based on intersecting tracks between a pedestrian and a neighboring vehicle), to an immature robot that cannot deal with the unexpected (not tracking a pedestrian deflected into its travel lane; forgetting that a pedestrian a few feet away is likely still somewhere close when tracking is lost), to a reasonable human driver (who could not have reacted that fast, even while saying a robot could have), and back to an immature robot that had no way to know the thing it saw in its camera and was driving over with ample sensory clues was in fact that same nearby untracked person. (See especially EXPR sections 3.2.3 and 3.3.4.)

**Off-nominal situations and safety.** At a higher level, the scenario points out that robotaxis will need to deal with severe off-nominal scenarios with potentially high consequences. Lacking a human safety driver, the opportunities for disastrously incorrect decisions in complex, messy situations based on incomplete sensor data in scenarios the design team feels are unforeseeable seem numerous.

Cruise internal discussion statements such as "the pedestrian is well past our lane of travel into the other lane" (QER 39) and "it would not be reasonable to expect that the other vehicle would speed up and proceed to hit the pedestrian, and then for the pedestrian to flip over the adjacent car and wind up in our lane" (QER 39) do not change the fact that the mishap occurred, and will not prevent future similar mishaps.

### 3.3   Potential lessons

Below are some potential lessons that could be helpful for Cruise and other designers of automated vehicle functions:
- High prediction confidence of pedestrian intent can be a risky basis for significant acceleration. Accelerating toward a vulnerable road user in the own-vehicle path reduces reaction time to surprise events, and can be contrary to rules of the road.
- A vulnerable road user mishap in an adjacent lane can present substantial risk due to inherently unpredictable outcomes, and should not be ignored.
- Arguments that superior sensor capability and fast reaction time will necessarily produce safer-than-human-driver outcomes overlook the more difficult areas of perception and prediction. Post-crash vulnerable road user detection can be expected to be especially challenging.
- Having another road user initiate a mishap does not absolve the AV from a responsibility to react in a reasonable way to inherently unpredictable events.
- Subjective judgements of the reasonableness of scenarios by developers should not override methodical safety engineering practices.

- Automating post-collision actions requires robust sensing and somewhat different prediction capabilities during and after collision.
- A so-called "Minimal Risk Condition" is, at best, a relative statement. At the very least, the position of all vulnerable road users – including potentially under the AV – should be unambiguous before any vehicle movement is attempted.

## 4     The immediate response

### 4.1     Immediate response timeline

1. (within +5 seconds) 9:29 PM: Remote Assistance at CRAC connected to AV, apparently with slightly delayed video and audio (QER at 21, 61). The pullover maneuver and pedestrian dragging was still in progress.
2. (+2 minutes) 9:31 PM: 911 emergency phone call dispatchers were alerted to the crash by a bystander calling in (QER 24). There is no indication that Cruise or their contractors ever contacted City emergency dispatchers.
3. (+3 minutes) 9:32 PM: The AV sent a 14-second video showing the collision to CRAC, but not the pullover maneuver/dragging segment (QER at 11).
4. (+8 minutes) 9:37 PM: City emergency responders are on scene. They ask CRAC via AV connection to keep AV in place pending extrication (QER 69).
5. (+10 to +15 minutes): Cruise Driverless Support Specialists ("DSS") arrive physically at the scene (QER 20).
6. (+20 minutes) 9:49 PM: CIRT labels accident "Sev-1" (minor collision). (QER 11) based on input from CRAC.[3]
7. (within +20 minutes): National pager system is activated to notify Cruise employees of a "minor" mishap (QER at 22).
8. (+45 minutes) 10:17 PM: Cruise contractors arrive on scene, noting post-crash AV movement and victim blood and skin patches on the ground (QER 11, 21).
9. (+2 hours) 11:31 PM: Cruise raises accident to "Sev-0" (major with injury) and initiates a War Room response, notifying more employees (QER 11).
10. (+2.5 hours) 11:55 PM: Cruise CEO joins the War Room Slack channel and shortly thereafter the War Room Google Meet (QER 22).
11. The QER cites a lack of "conclusive evidence" that Cruise employees including senior leadership had knowledge of the pedestrian dragging on Oct. 2 (QER at 23). Some participants recall a discussion about the pedestrian having been dragged before a shift end at 4:00 AM Oct 3 (QER 22 fn 21).

---

[3] The Sev-1 characterization by Cruise is based on "the fact that Cruise's Remote Assistance had characterized the accident a 'minor collision…'" (QER 41). On the other hand, "there is no indication Cruise spoke with the Remote Assistance contractors who were interacting with the AV" until weeks later (QER 103). Perhaps there was messaging rather than voice communications that are not described. In any event, communications and entries in any communication records between CRAC and Cruise seem significantly under-reported in the QER.

### 4.2    Post-crash analysis

**What went right:** The remote assistance channel opened essentially immediately, giving CRAC the ability to see and hear what had happened, including both the AV/pedestrian collision and the dragging sequence. This channel was successfully used to communicate with on-scene responders who, for example, advised CRAC to keep the vehicle stationary. A CIRT initial response occurred relatively quickly after the mishap, providing Cruise with the potential to coordinate a company-wide response.

**Failure to stay on plan:** There was a response playbook, but it was not followed because it was "too manually intensive" (QER 22). Although the War Room was supposed to address accident causation and next steps, there was a quick focus on media narrative "almost exclusively" (QER 23). CIRT arguably did not know about the dragging, and was concerned that the media and an SFFD statement emphasized a pedestrian pinned under the vehicle rather than the initial Nissan collision (QER 24). While this reaction might be attributable to the equivalent of an inevitable Fog of War, deviating from the procedural playbook was a missed opportunity to stay on plan.

**Failure to request timely remote intervention:** One of the challenges of immature vehicle automation is ensuring that the AV asks for help when it needs to instead of doing something dangerous without additional support. The AV in this case got that decision wrong. The connection to CRAC was established quickly, but the AV had already initiated the pullover maneuver that undoubtedly led to a severe increase in pedestrian injury, rather than waiting for CRAC guidance.

One can imagine a series of design decisions that lead to such an outcome, especially motivated by intense scrutiny and criticism of previous robotaxi strandings blocking traffic in highly publicized events [5]. Nonetheless, designing a vehicle that all on its own decides to move after striking an unclassified object (presumed to be a vulnerable road user (EXPR 32)) seems a poor system-level safety approach.

**Failure to alert emergency responders.** There is no indication that Cruise or its contractors initiated contact to City emergency services. Similarly, there is no indication CRAC or any other Cruise operational team considered emergency service contact. The first mention of emergency responders is when they were already at the scene contacting CRAC via physically accessing the vehicle communications feature to ask for the vehicle to remain stationary. By all accounts, CRAC knew a pedestrian injury had occurred. CRAC should have reached out to City emergency responders immediately.

**Failure of coordination between operations and engineering/management:** CRAC and on-the-ground response teams both knew a serious pedestrian injury had occurred. The initial misclassification as Sev-1 is unexplained. According to the QER, CIRT did not realize a pedestrian dragging had occurred in a timely manner. A failure to coordinate the CIRT response with CRAC at the time of an incident, and especially when the incident was upgraded to Sev-0 (major) seems a major process failure. (See [2] for additional perspective.)

**Premature removal of safety supervisor:** A physically present safety supervisor could have prevented the pedestrian dragging, and arguably initiated a safety stop well before the first pedestrian impact based on a developing high-risk situation. Indeed, many other incidents such as mass AV strandings would likely have been mitigated or avoided entirely by a physically present backup driver in the AV. Removal of the safety supervisor seems motivated by business and publicity interests rather than a primary focus on safety. There were ample warnings that safety supervisors had been removed prematurely in the form of numerous incidents of emergency responder interference.

### 4.3   Potential lessons

- While quick movement of an AV out of travel lanes is desirable, doing so after a detected collision seems high risk due to the high variability and uncertainty of possible post-collision scenarios. A brief wait for remote assistance seems justified by safety considerations regardless of public pressure to minimize road disruptions.
- It is difficult to manage a crisis in general. Deviating from procedures is all too easy, but makes things worse. Periodic practice of crisis responses is required for effective execution when the real thing happens. Such training is not mentioned.
- Remote assistance operators should routinely and immediately alert emergency responders if they have any reason to believe an injury collision has or might have occurred. In this incident, Cruise got lucky that a bystander made that call.
- There is no explanation given for the apparent communication gap between CRAC and CIRT. Incident response communications should be scripted in the response plan, with specific handoffs and information to be transferred. CRAC should keep a contemporaneous log of such communications. Potential bad news, including especially road user injuries, should be highlighted to the CIRT and logged.
- There is no mention of a methodical approach to preserve evidence such as the vehicle configuration, communication logs, and other technical data beyond doing a data download from the vehicle. Post-incident procedures should address this area.
- In retrospect, the human in-vehicle safety supervisor for Cruise vehicles was likely removed prematurely. Having one might have prevented the initial AV pedestrian collision by reducing vehicle speed in response to a dangerous road situation. Even if that had not happened, an in-vehicle safety supervisor could be expected to notice and intervene with running over a pedestrian followed by dragging, and would be a backup method for contacting emergency services.

## 5   Organizational Response

We treat everything happening starting the next day (October 3, 2023) as the organizational response. CMT activity lasted until approximately 24 hours after the crash, but changed in tenor to a public relations and regulatory response for these activities within a few hours of the crash. Here is a timeline, abbreviated due to space constraints:
1. (+6.3 hours) 3:45 AM: A War Room Slack message unambiguously communicates the AV pulled forward with the pedestrian under the vehicle. This was done via CMT accessing data from CRAC (QER 27) rather than via a proactive hand-off process from CRAC to CMT. The QER says that no other "Cruise employee" had accessed this data previously (QER 28), but we note that CRAC staff are contractors, so the QER statement can easily be misunderstood.

Anatomy of a Robotaxi Crash: The Cruise Pedestrian Dragging Mishap    132. Oct. 3: Cruise meets with DMV. DMV claims that Cruise did not disclose the pullover maneuver and dragging, only learning about it from another government agency later (QER 1-2). Internet connectivity issues and lack of affirmative discussion regarding the dragging are said to have contributed to the situation (QER 2). Interviews with DMV personnel were out of scope for the QER.
3. Oct. 3: NHTSA was said to have received the full video (QER 2). However, NHTSA was not shown the full video during the meeting for various reasons and the pedestrian dragging was not discussed (QER 41-42).
4. Oct. 3: City officials (SF MTA, SFPD, SFFD) shown the full video with discussion about the pedestrian dragging; no apparent internet issues (QER 2-3).
5. Oct. 3: Media outlets were shown a video that ended at the moment of impact and omitted the pedestrian dragging (QER 2). Cruise leadership believed they did not have an obligation to disclose the full details to the press (QER 3).
6. Oct. 3: A required "1-Day Report" to NHTSA failed to mention the pedestrian dragging (QER 15). This is said to have been due to a paralegal with "little oversight" drafting and filing (QER 98). However, Slack messages document Cruise's Deputy General Counsel and Communications Director both affirmatively approving a draft narrative missing that information (QER 49).
7. Oct. 11: A 10-Day report to NHTSA filed by the same paralegal also omits mention of the pedestrian dragging (QER 16).
8. Oct. 13-16: Cruise meets with DMV and shares full video (QER 69).
9. Oct. 24: DMV issues suspension order of Cruise's California operating permits. Cruise suspends California driverless operations (QER 87-88).
10. Nov. 2: Cruise submits a 30-Day report to NHTSA that includes mention of pedestrian dragging, along with a recall of 950 vehicles (QER 88).
11. Aftermath: Eleven Cruise employees involved with briefings to government regulators depart, including resignation of the CEO, followed by a 24% Reduction in Force (QER 91).

### 5.1 Organizational response analysis

**What went well:** Cruise proactively contacted local government and regulators to set up meetings. A War Room activation plan was carried out, ensuring that relevant Cruise stakeholders had a means of communication. Videos of the crash were created, including a video of the full mishap sequence, before external stakeholder meetings.

**Narrow focus on narrative:** The Cruise management team immediately became focused on asserting control of a public narrative they saw as unfair and felt "under siege" (QER 24,26). In part, this was apparently due to incomplete knowledge on both sides. For example, SFFD was making statements based on having been on the scene in which grievous injuries to a pedestrian had occurred, but without first-hand information about the initial Nissan impact. On the other hand, Cruise knew about the Nissan impact, but apparently did not know about the pedestrian dragging when establishing their public narrative, with that momentum carried forward into the following day.

**Failure to stand down fleet:** The Cruise decision not to stand down fleet operations after a dramatic, severe pedestrian injury was justified by the CEO labeling the event an "extremely rare event" "edge case" and considering the event in the context of "the overall driving and safety records" (QER 35).

**Failure To Proactively Disclose Material Bad News:** By the time regulatory and media discussions were happening, more than 100 people in Cruise, including the CEO, knew of the pedestrian dragging (QER 28). Nonetheless, this fact was not proactively disclosed, even though Cruise knew it was their "biggest issue" (QER 40).

Regardless of intentions, later revelations/clarifications that pedestrian dragging had been under-disclosed caused a furor (e.g., QER 73). Arguably this communication mistake (rather than the mishap itself) was the main reason Cruise operations eventually had to be shut down. A common industry attitude is not volunteering information not explicitly asked for by government regulators. That attitude might avoid initiating negative news cycles, but ultimately resulted in a dramatically bad outcome for Cruise.

### 5.2    Potential lessons

Ultimately, as the saying goes, the cover-up is always worse.
- While it is natural to be defensive after a mishap involving an organization's technology, lack of transparency and lack of affirmative disclosure of bad news can bring serious negative consequences, including loss of trust and stakeholder backlash. A key misstep was deciding to double down on not proactively revealing the pedestrian dragging in various ways. This included not ensuring regulators noticed the pedestrian dragging issue, and not correcting overly favorable media reports.
- The fear of triggering a negative media cycle (QER 31) ultimately resulted in increased harm to the company. If it becomes clear that stakeholders have incorrect or inadequate information (1) acknowledge as quickly as possible to those stakeholders that a revision will be necessary and (2) use extra effort to disclose fully and completely what is known to the organization as soon as possible.
- Do a safety stand down in response to a major event. This could have given Cruise more credibility with regulators, and blunted the public perception of their emphasis on scaling up operations regardless of cost or harm.
- Organizations should consider having a communications specialist who is not invested in the organization's narrative coordinate crisis communications to avoid narrative capture by stakeholders heavily incentivized to continue operations.

## 6    Conclusions

Another major mishap involving an AV is inevitable. Road travel is not perfectly safe, and such a mishap occurring should not all on its own be a reason to discontinue development of AV technology. However, the wrong mishap occurring in the wrong way, which is handled poorly, can potentially pose an existential threat to a company – and perhaps to the entire industry. This Cruise mishap has unquestionably reverberated throughout the industry, in large part because of the mishandled regulatory response.

Technical issues related to the crash identified include weaknesses in recognizing and responding to nearby mishaps, challenges in building an accurate world model of a post-collision scenario, and the inadequacy of a so-called "minimal risk condition" strategy in complex situations. Post-collision issues center around poor organizational discipline in responding to a mishap as well as overly aggressive post-collision automation that made a bad situation worse. Organizational issues identified center around a reluctance to admit to and address a mishap, ultimately causing organizational harm.

Cruise had previously spent significant effort to spin the narrative as a bad human driver in the Nissan causing a mishap, with their robotaxi doing the best it could in a situation nobody could have reasonably predicted. However, casting blame on the driver who initially hit the pedestrian (even if deserved) does not serve the purpose of identifying safety issues for the robotaxi that should be improved.

With the availability of the QER and EXPR, we have the opportunity to learn more about other aspects of the mishap and its aftermath, with key lessons listed in sections 3.3, 4.3, and 5.2. We present potential lessons regarding the vehicle's handling of a pre-crash scenario and post-crash scenario. We also present potential lessons regarding both the immediate and longer-term organizational responses to such a mishap.

Rather than attempting to create an accident report to parallel the one presented by the EXPR, this paper seeks to make available information in the reports more accessible to a broader audience. A potential threat to the validity of these findings stems from incomplete information in the reports. The QER is based on interviews limited to available employees and contractors but not outsiders (QER 4-5), and the EXPR is heavily redacted in places. We optimistically assume that redacted information would not undermine or contradict visible information. Nonetheless, given the situation we recommend assuming the information presented in both the QER and EXPR are crafted to present the most favorable picture possible toward Cruise's stated goal of regaining regulatory and public trust.

No external support funded the preparation of this paper.